# Dynamic Trajectory Adaptation for Efficient UAV Inspections of Wind Energy Units

Serhii Svystun
*Department of Computer Engineering and Information Systems
of Khmelnytskyi National University*
Khmelnytskyi, Ukraine
svystuns@khmnu.edu.ua

Oleksandr Melnychenko
*Department of Computer Engineering and Information Systems
of Khmelnytskyi National University*
Khmelnytskyi, Ukraine
melnychenko@khmnu.edu.ua

Pavlo Radiuk
*Department of Computer Science
of Khmelnytskyi National University*
Khmelnytskyi, Ukraine
radiukp@khmnu.edu.ua

Oleg Savenko
*Department of Computer Engineering and Information Systems
of Khmelnytskyi National University*
Khmelnytskyi, Ukraine
savenko_oleg_st@ukr.net

Anatoliy Sachenko
*Research Institute for Intelligent Computer Systems
of West Ukrainian National University*
Ternopil, Ukraine
*Department of Teleinformatics of Kazimierz Pulaski University of Radom*
Radom, Poland
as@wunu.edu.ua

Andrii Lysyi
*Department of Computer Engineering and Information Systems
of Khmelnytskyi National University*
Khmelnytskyi, Ukraine
andriilysyi@khmnu.edu.ua

*Abstract*—The research presents an automated method for determining the trajectory of an unmanned aerial vehicle (UAV) for wind turbine inspection. The proposed method enables efficient data collection from multiple wind installations using UAV optical sensors, considering the spatial positioning of blades and other components of the wind energy installation. It includes component segmentation of the wind energy unit (WEU), determination of the blade pitch angle, and generation of optimal flight trajectories, considering safe distances and optimal viewing angles. The results of computational experiments have demonstrated the advantage of the proposed method in monitoring WEU, achieving a 78% reduction in inspection time, a 17% decrease in total trajectory length, and a 6% increase in average blade surface coverage compared to traditional methods. Furthermore, the process minimizes the average deviation from the optimal trajectory by 68%, indicating its high accuracy and ability to compensate for external influences.

I. INTRODUCTION

Over the past decades, wind turbines have become essential components of the global shift towards sustainable energy. At the same time, various components of wind turbines, i.e., WEUs, require constant and timely inspections to maintain their efficiency and safety [1, 2]. Traditionally, these inspections are conducted manually by skilled technicians who physically assess the turbines [3, 4], often requiring them to climb the massive structures or use binoculars and cameras from the ground. This approach, while thorough, is fraught with challenges [5]. The physical strain on inspectors, the time-consuming nature of the task, and the risk of human error all contribute to the inefficiency of manual inspections [6]. Additionally, harsh weather conditions, such as strong winds and rain, can make these inspections both problematic and dangerous [7], potentially compromising the health and safety of the personnel involved.

Given these challenges, there is a compelling case for hypothesizing that the issues associated with human inspection of wind energy units can be efficiently addressed using UAVs combined with automated trajectory planning. UAVs, equipped with advanced optical and thermal sensors [8, 9], can be programmed to follow precise flight paths around the wind turbines [10], capturing high-resolution images and data that can be analyzed for defects and wear [11]. The automation of these flight trajectories allows for consistent and repeatable inspections, minimizing the variability introduced by human operators. Furthermore, UAVs can operate in conditions that would be too dangerous or impractical for humans [12, 13], ensuring that inspections can be conducted more often and with higher accuracy [14].

The increasing need for efficient and safe wind turbine inspections, driven by the growing role of renewable energy, highlights the limitations of traditional manual methods. These methods are time-consuming [15], risky for human operators [16], and prone to errors due to harsh environmental conditions. The use of UAVs offers a promising alternative for automated inspections [17], ensuring more frequent, accurate, and safer maintenance procedures [18]. This study aims to address challenges in dynamic trajectory adaptation and efficient data collection, reducing inspection time and improving surface coverage of turbine components. The main contributions of this research are as follows:

- Developed an automated UAV trajectory planning method that reduces inspection time by 78%.
- Improved blade surface coverage by 6% through real-time adaptation to blade positioning.
- Minimized deviations from optimal flight paths by 68% using a proportional–integral–derivative controller to compensate for external influences like wind.

The paper reviews related work on wind turbine inspection (Section II), presents the proposed automated trajectory method (Section III), analyzes experimental results (Section IV), and concludes with a summary and future work (Section V).

## II. RELATED WORKS

The field of UAV-based wind turbine inspection has evolved, driven by the need for more efficient alternatives to traditional methods. Research has focused on automated trajectory planning, defect detection, and advanced data processing [19, 20]. Car et al. [21] proposed a spiral-based trajectory algorithm to improve blade coverage, while Pérez et al. [22] generated rectangular trajectories based on blade geometry. Nonetheless, both approaches exhibit limitations in dynamically adapting to real-time changes in blade positions during flight, which is a critical aspect of adequate UAV-based inspections.

Advancements in defect detection have also been notable. Shihavuddin et al. [23] employed deep learning techniques to automatically detect surface cracks and other damage on turbine blades, showcasing significant progress in the field. Memari et al. [24] extended this work by incorporating thermal imaging techniques to identify internal defects and temperature anomalies, adding another layer of precision to the inspection process. Moreover, Yang et al. [9] advanced the integration of multi-sensor data, combining optical and thermal sensors to improve the accuracy of defect characterization. Gohar et al. [25] enhanced defect detection through efficient image slicing, which is particularly effective in handling ultra-high-resolution images. However, Gohar's solution relies heavily on manual UAV control, restricting its scalability and efficiency.

Traditionally, UAV control for wind turbine inspections has been conducted manually, with operators independently planning and adjusting flight trajectories in real-time [26, 27]. As the demand for more autonomous solutions grows, recent studies have introduced various approaches to automated path planning [28]. Castelar Wembers et al. [29] presented a 2D-LiDAR sensor-based approach that reduces inspection time. Despite its innovative nature, this approach faces challenges in scalability during real-world applications. Li et al. [30] effectively combined UAV path planning with defect detection, but their method does not adequately address the variability of environmental conditions or the limitations of current sensor technologies. Likewise, Liu et al. [31] proposed a two-stage heuristic algorithm aimed at optimizing trajectory quality. However, this technique lacks real-time adaptability, which is crucial for responding to dynamic inspection conditions.

Therefore, this study's aim is to improve UAV flight trajectory planning for wind turbine inspections by developing a novel automated method that accounts for real-time component positioning and external factors like wind. To achieve this aim, the following tasks are undertaken:

- Design an algorithm for efficient flight path planning based on turbine component positioning.
- Implement real-time trajectory adjustments to account for blade movement and wind.
- Evaluate the method's performance through experiments comparing it to traditional approaches.

## III. METHOD

The *input data* for the method is a set of coordinates of wind turbines in a three-dimensional environment. The defined spatial zone of wind turbines $Z_i$ is represented as a sphere with center $C_i$ and radius $R_i$. This model clearly defines the boundaries of the space in which the system is operated, thereby ensuring data collection using UAVs. The proposed method is divided into two blocks based on its functionality and schematically illustrated in Fig. 1.

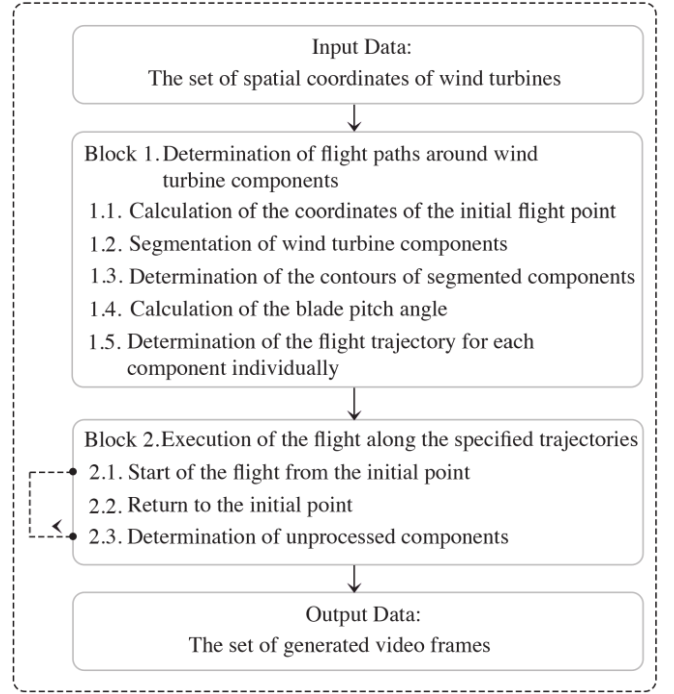

Fig. 1. The scheme of the proposed method reflects the UAV-based wind turbine inspection process, from input data of spatial coordinates through flight path determination and execution, resulting in video frames.

*Block 1*. Determination of flight trajectories around wind turbine components.

*Step 1.1.* Determination of the initial flight point. The efficiency of data collection from WEUs using UAVs significantly depends on the selection of the initial flight point. Considering the variability in the spatial positioning of wind turbine blades, choosing a point that provides an optimal view of all WEU components for subsequent trajectory planning is essential. In the proposed method, the initial point $S_p$ is determined based on the analysis of the spatial zone of the wind turbine $Z_i$, which is modeled as a sphere with center $C_i$ and radius $R_i$:

$$Z_i = (x, y, z) \in \mathbb{R}^3 \mid (x - x_i)^2 + (y - y_i)^2 + (z - z_i)^2 \leq R_i^2, \quad (1)$$

where $(x_i, y_i, z_i)$ are the coordinates of the sphere's center.

The scheme of initial point $S_p$ is illustrated in Fig. 2.

The center of the sphere $C_i$ is calculated as the average value of the coordinates of all the wind turbine blades:

$$C_i = \left( \frac{\sum_{k=1}^{n} x_k}{n}, \frac{\sum_{k=1}^{n} y_k}{n}, \frac{\sum_{k=1}^{n} z_k}{n} \right), \quad (2)$$

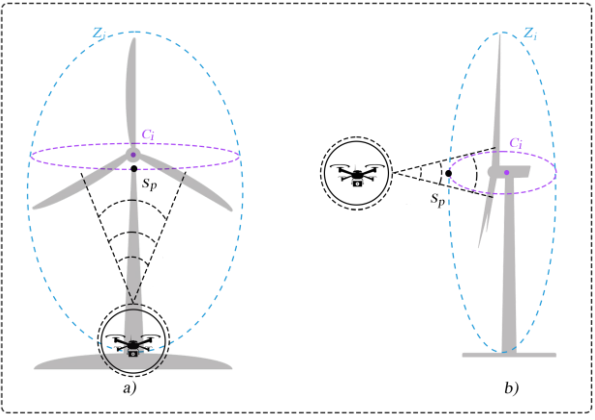

Fig. 2. The spatial positioning and trajectory planning for a UAV inspecting a wind turbine with two perspectives: (a) a frontal view with the UAV determining its initial flight point $S_p$ within the inspection zone $Z_i$, and (b) a side view emphasizing the calculated flight trajectory around the turbine blades, highlighting the optimized path $C_i$ for component coverage.

where $n$ represents the number of blades, and $(x_k, y_k, z_k)$ are the coordinates of the wingspan (edge tips) of each blade.

The radius of the viewing space $R_i$ in (1) is determined as the maximum distance from the center $C_i$ by (2) to the endpoints of the blades:

$$R_i = \max\left\{\sqrt{(x_k - x_i)^2 + (y_k - y_i)^2 + (z_k - z_i)^2}\right\}_{k=1}^{n}. \quad (3)$$

To ensure targeted data collection and enable the initial analysis of the WEU, the initial point $S_p$ is selected opposite the drive mechanisms (Fig. 3a) at a distance $R_i$ calculated by (3) from the center $C_i$: $S_p = (x_i + R_i, y_i, z_i)$. The output data from step 1.1 are the coordinates of the initial point $S_p$, which will be used for further UAV trajectory planning.

*Step 1.2.* Segmentation of WEU components. The positioning of the UAV at the coordinate point $S_p$ provides the necessary view for conducting component segmentation of the WEU, facilitated by the acquired visual image (Fig. 3b). Upon reaching the initial point $S_p$, the optical sensor generates an image $I_m$ of the wind energy unit, which captures all the essential components of the WEU (Fig. 3c).

For further analysis and the formation of the flight trajectory, it is necessary to isolate individual WEU components in the image: blades, tower, and nacelle. To segment the image $I_m$, the software tool Detectron2 [32] is used with the function $f_{predictor}$. When applied to the image $I_m$, this function returns a set of segments: $S_m = s_1, s_2, \ldots, s_m$, where $s_i \subseteq I_m$ represents the subset of pixels belonging to the $i$-th segment. For each segment $s_i$, a mask $M_i$ and bounding box $B_i$ are defined, indicating its position and shape on the image. After segmentation, the background of the image is eliminated using the function $f_{bitwise}$ [33] from the OpenCV library [34]. As a result, we obtain a transformed image $I_t$, which contains only the segmented WEU components.

*Step 1.3.* Determination and filtration of WEU component contours. Based on the obtained masks $M_i$, binarized masks $M_{b,i}$ are created for each segment:

$$M_{b,i}(x,y) = \begin{cases} 1, & \text{if } M_i(x,y) = 1; \\ 0, & \text{else } M_i(x,y) \neq 1. \end{cases} \quad (4)$$

Binarized masks calculated by (4) are used to identify the contours of WEU components using the function $f_{findContours}$ [35] from the OpenCV library. As a result, we obtain a set of contours $C_m$:

$$C_m = \bigcup_{i=1}^{m} C_i, \quad (5)$$

where $C_i$ is the set of contours identified in the mask $M_{b,i}$. To remove noise and irrelevant contours from $C_m$ calculated by (5), an area-based filter is applied:

$$C_{filtered} = c_i \in C_m | A(c_i) > A_{threshold}, \quad (6)$$

where $A(c_i)$ is the area of the contour $c_i$, and $A_{threshold}$ is the threshold area value. This process allows for a clear separation of independent objects in the image (Fig. 4d).

The output of this step is the set $C_{filtered}$ calculated by (6), which contains the filtered contours corresponding to the components of the wind energy unit.

*Step 1.4.* Determination of the blade pitch angle. For each contour $c_i$ from the set $C_m$, corresponding to a wind turbine blade, the pitch angle $\theta_i$ determined. To achieve this, the minimum bounding rectangle $rect_i$ is computed using the function *minAreaRect*. The blade pitch angle is defined as the angle of inclination of the line passing through the top and bottom points of this rectangle:

$$\theta_i = \arctan\left(\frac{y_{bottom,i} - y_{top,i}}{x_{bottom,i} - x_{top,i}}\right), \quad (7)$$

where $(x_{top,i}, y_{top,i})$ and $(x_{bottom,i}, y_{bottom,i})$ are the coordinates of the top and bottom points of the rectangle. Based on the value of the angle $\theta_i$, the classification of the blade's pitch type is illustrated in Fig. 4, and includes the following categories:

1. Acute tilt (Fig. 4a): $\theta_i \in (30°, 60°] \cup [120°, 150°)$.

2. Horizontal tilt (Fig. 4b): $\theta_i \in [0°, 30°] \cup [150°, 180°]$.

3. Vertical tilt (Fig. 4c): $\theta_i \in (60°, 120°)$.

The output data of this section consists of the set of values determined for each individual blade, which we represent as $\Theta_i = \{\theta_i^1, \theta_i^2, \theta_i^3\}$ for each $i$-th blade.

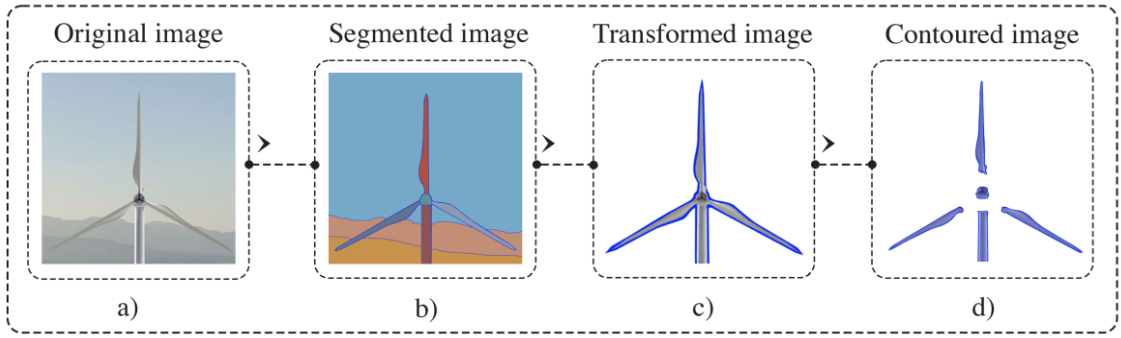

Fig. 3. The sequential stages of image processing in the proposed method: (a) the original captured image, (b) the segmented image highlighting individual components, (c) the transformed image isolating relevant features, and (d) the final contoured image used for detailed analysis and trajectory planning.

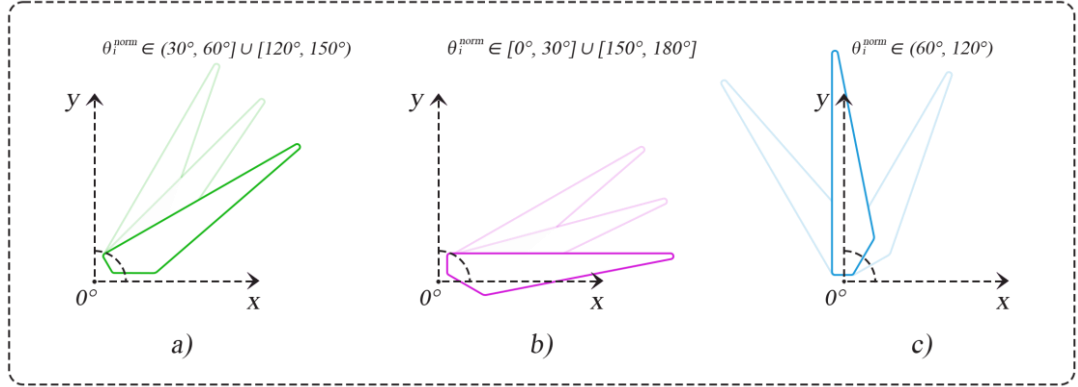

Fig. 4. The classification of wind turbine blade tilt types based on their angular orientation in the x-y plane: (a) acute tilt with angles ranging from 30° to 60° and from 120° to 150°, (b) horizontal tilt with angles from 0° to 30° and from 150° to 180°, and (c) vertical tilt with angles from 60° to 120°, demonstrating the variability in blade positioning for optimized inspection and analysis.

*Step 1.5.* Formation of flight trajectories. For the inspection of blades, dynamic trajectories $S_i$ are formed, which adapt to their tilt angle $\theta_i$ in real time:

$$S_i(t) = \begin{cases} f_{\text{path}}(S_{p_i}, t), & t \in [0, T_{\text{fb}}]; \\ S_{\text{fb},i}(T_{\text{fb}}) + (S_{p_i} - S_{\text{fb},i}(T_{\text{fb}})) \cdot \\ \left(\dfrac{t - T_{\text{fb}}}{T_{\text{cb}}}\right), & t \in [T_{\text{fb}}, T_{\text{fb}} + T_{\text{cb}}], \end{cases} \quad (8)$$

where $S_{p,i}$ is the initial flight point of the *i*-th blade; $T_{\text{fb}}$ is the flight time; $T_{\text{cb}}$ is the return time to the initial point; $f_{\text{fb},i}(S_{p,i}, \theta_i, t)$ is the function that describes the flight trajectory around the *i*-th blade, taking into account its tilt angle $\theta_i$; $f_{\text{cm},i}(S_{\text{fb},i}(T_{\text{fb}}), S_{p,i}, t)$ is the function that describes the UAV's return trajectory from the final point of the flight $S_{\text{fb},i}(T_{\text{fb}})$ to the initial point $S_{p,i}$.

The set of output trajectories from (8) for all *N* WEU components is denoted as:

$$T = S_i(t) | i \in 1, 2, \ldots, N. \quad (9)$$

The output of this step is the set of trajectories *T* as presented in (9), which is used for automated UAV control during the flight around the WEU. Fig. 5 presents the scheme of the developed method for determining the flight trajectories of WEU components.

It is important to note that external factors such as wind and turbulence significantly affect the UAV's flight trajectory. To compensate for this influence and ensure flight stability, the proposed method utilizes a proportional–integral–derivative (PID) controller algorithm [36]. The correction coefficient vectors for each coordinate are determined by the following system of equations:

$$\begin{cases} u_x(t) = K_P e_x(t) + K_I \int_0^t e_x(\tau) d\tau + K_D \dfrac{d}{dt} e_x(t); \\ u_y(t) = K_P e_y(t) + K_I \int_0^t e_y(\tau) d\tau + K_D \dfrac{d}{dt} e_y(t); \\ u_z(t) = K_P e_z(t) + K_I \int_0^t e_z(\tau) d\tau + K_D \dfrac{d}{dt} e_z(t), \end{cases} \quad (10)$$

Unlike the blades, the tower and nacelle of the WEU are static elements, so fixed trajectories can be used for their inspection, which are determined in advance. These trajectories calculated by (10) are formed considering the geometric parameters of the WEU and the optimal viewing angles to obtain high-quality images.

## IV. RESULTS AND DISCUSSION

The data in Table 1 compares the efficiency of two UAV control methods for wind turbine inspection across various scenarios. These scenarios simulate different inspection conditions, such as the number of turbines, terrain characteristics, and weather conditions.

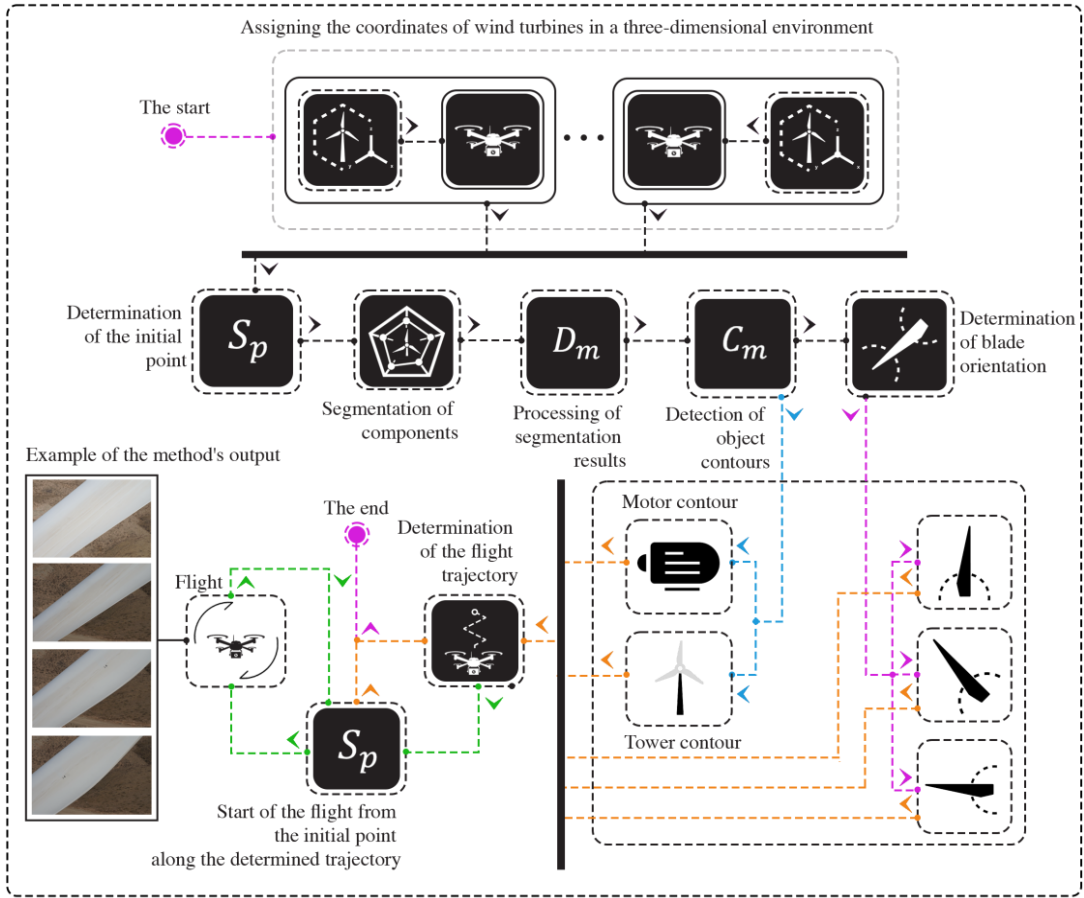

Fig. 5. This figure details the proposed method for UAV-based wind turbine inspection, highlighting the stages from assigning turbine coordinates, component segmentation, and contour detection to determining flight trajectory and blade orientation.

As shown in Table 1, the proposed automated method demonstrates significant advantages across all considered scenarios, reducing inspection time, shortening trajectory lengths, and improving blade surface coverage. The use of a PID algorithm to compensate for external influences, such as wind and turbulence, plays a crucial role in achieving these results.

Fig. 6 presents a comparison of traditional and automated UAV control methods across several performance metrics. The figure compares the time spent on inspecting wind turbines across different scenarios for both methods, displays the total distance covered by UAVs during the inspection, illustrates the percentage of blade surface covered by each technique, and shows the deviation from the ideal path for both control methods. As evident from the presented data, the proposed automated method demonstrates significant advantages, reducing the overall inspection time by 78%, decreasing the total trajectory length by 17%, increasing the average blade surface coverage by 6%, and minimizing the average deviation from the optimal trajectory by 68%. The application of the PID algorithm for compensating external influences, such as wind and turbulence, plays a crucial role in achieving these results.

The method also minimizes deviations from the optimal trajectory, indicating high accuracy and the ability to counteract external influences. A significant advantage is the elimination of the need for operator involvement, which reduces costs and minimizes the risk of errors. Similar improvements in inspection efficiency have been demonstrated by our previous work [11], in which we implemented an automated UAV control system capable of real-time adaptation.

TABLE I. COMPARISON OF UAV CONTROL METHODS EFFICIENCY FOR WIND TURBINE (WT) INSPECTION

| Scenario | Control method: manual/automated | Total inspection time (min) | Total trajectory length (m) | Average blade surface coverage (%) | Average deviation from optimal trajectory (m) | Number of operators |
|---|---|---|---|---|---|---|
| Three WTs, open terrain, weak wind (up to 5 m/s) | Manual, 1 UAV | 90 | 1400 | 88 | 3.0 | 3 |
|  | Proposed automated, 3 UAVs | 8 | 1100 | 95 | 1.0 | 0 |
| One WT, complex terrain, strong wind (8-12 m/s) | Manual | 35 | 600 | 82 | 5.0 | 1 |
|  | Proposed automated | 7 | 480 | 92 | 1.5 | 0 |
| Two WTs, different heights, medium wind speed (5-8 m/s) | Manual | 50 | 1000 | 86 | 3.5 | 2 |
|  | Proposed automated | 9 | 800 | 94 | 1.2 | 0 |
| Five WTs, open terrain, no wind | Manual | 150 | 2300 | 89 | 2.5 | 5 |
|  | Proposed automated | 12 | 1900 | 96 | 0.6 | 0 |

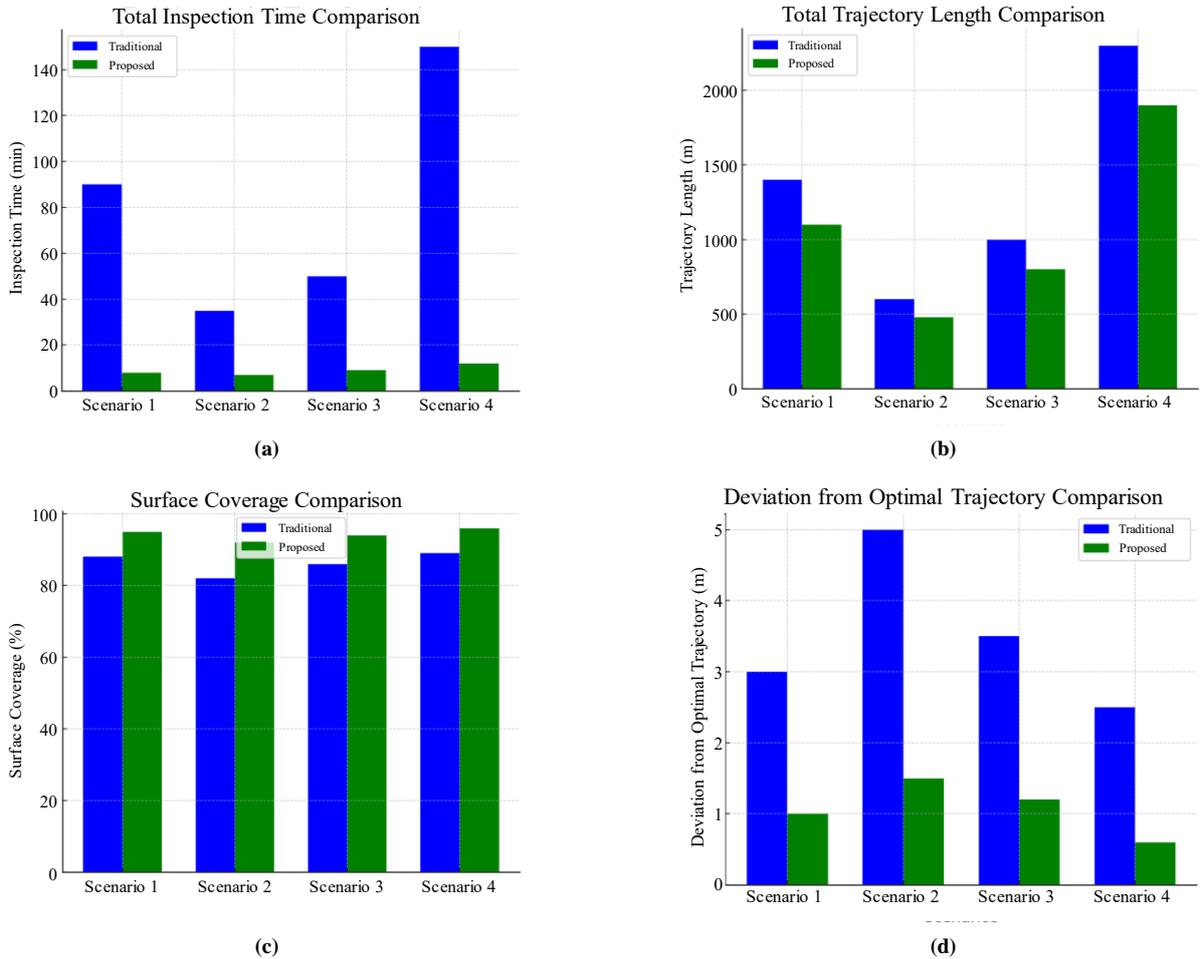

Fig. 6. This figure compares traditional (blue) and proposed (green) UAV-based wind turbine inspection methods across four scenarios, highlighting total inspection time (a), trajectory length (b), surface coverage (c), and deviation from optimal trajectory (d). The proposed method consistently outperforms the traditional approach, showing reduced inspection times, shorter trajectory lengths, improved surface coverage, and minimized deviations from optimal trajectories, demonstrating its efficiency and accuracy in diverse conditions.

Despite these significant advantages above, the proposed method has some limitations. Firstly, accurate data about the geometry and location of the WEUs is essential for the correct operation of the process. Furthermore, like any complex system, the automated UAV control system can malfunction, which could lead to an emergency. Finally, the implementation and operation of the automated system requires experts in robotics, programming, and wind energy.

## V. Conclusion

In this work, we introduced a novel automated method for determining UAV flight trajectories for WEU inspections. The method is based on pre-calculated and optimized flight trajectories. It accounts for wind turbines' static geometry and positioning and allows for incorporating external factors like wind, which should be addressed in other studies. By integrating these elements, the proposed approach significantly enhances the automation of the inspection process, resulting in reduced inspection time and improved data accuracy. In particular, the authors' contribution effectively reduces the overall inspection time by 78%, decreases the total trajectory length by 17%, and increases the average blade surface coverage by 6%. Moreover, the average deviation from the optimal trajectory is minimized by 68%, showcasing the high precision and reliability of the proposed approach. These improvements result in significant cost and time savings, making the inspection process more efficient and less dependent on human intervention. Despite these promising results, the proposed method is limited by its reliance on accurate input data and the potential for system malfunctions, requiring proper implementation expertise.

The following research phase will focus on implementing Block 2, responsible for executing the flight around the WEU along the specified trajectories. That will involve developing algorithms for UAV control, considering flight dynamics and external factors such as wind and turbulence, and generating precise trajectories for inspecting dynamic components like blades and static components, such as the tower and nacelle.

## VI. References

[1] R. Wiser *et al.*, "Land-based wind market report: 2023 Edition", Office of Scientific and Technical Information (OSTI), Oak Ridge, 1996790, Aug. 2023. Accessed: Aug. 25, 2024. [Online]. Available: https://doi.org/10.2172/1996790

[2] E. A. Virtanen *et al.*, "Balancing profitability of energy production, societal impacts and biodiversity in offshore wind farm design," *Renewable Sustain. Energy Rev*., vol. 158, p. 112087, Apr. 2022, doi: 10.1016/j.rser.2022.112087.

[3] W. Chen, Y. Qiu, Y. Feng, Y. Li, and A. Kusiak, "Diagnosis of wind turbine faults with transfer learning algorithms," *Renew. Energy*, vol. 163, pp. 2053–2067, Jan. 2021, doi: 10.1016/j.renene.2020.10.121.

[4] T. Sun, G. Yu, M. Gao, L. Zhao, C. Bai, and W. Yang, "Fault diagnosis methods based on machine learning and its applications for wind turbines: A review," *IEEE Access*, vol. 9, pp. 147481–147511, 2021, doi: 10.1109/ACCESS.2021.3124025.

[5] W. Yang, P. J. Tavner, C. J. Crabtree, Y. Feng, and Y. Qiu, "Wind turbine condition monitoring: technical and commercial challenges," *Wind Energy*, vol. 17, no. 5, pp. 673–693, 2014, doi: 10.1002/we.1508.

[6] Y. Sun *et al.*, "UAV and IoT-based systems for the monitoring of industrial facilities using digital twins: Methodology, reliability models, and application," *Sensors*, vol. 22, no. 17, Art. no. 17, 2022, doi: 10.3390/s22176444.

[7] S. Roga, S. Bardhan, Y. Kumar, and S. K. Dubey, "Recent technology and challenges of wind energy generation: A review," *Sustain. Energy Technol. Assessments*, vol. 52, no. C, p. 102239, Aug. 2022, doi: 10.1016/j.seta.2022.102239.

[8] I. Paliy, A. Sachenko, V. Koval, and Y. Kurylyak, "Approach to face recognition using neural networks," in *2005 IEEE Int. Conf. Intell. Data Acquisition Adv. Comput. Syst.: Technol. Appl. (IDAACS)*, Sofia, Bulgaria, Sep. 5–7, 2005. New York, NY, USA: IEEE, 2005, pp. 112-115. doi: 10.1109/IDAACS.2005.282951.

[9] C. Yang, X. Liu, H. Zhou, Y. Ke, and J. See, "Towards accurate image stitching for drone-based wind turbine blade inspection," *Renew. Energy*, vol. 203, pp. 267–279, Feb. 2023, doi: 10.1016/j.renene.2022.12.063.

[10] Z. Zhang and Z. Shu, "Unmanned aerial vehicle (UAV)-assisted damage detection of wind turbine blades: A review," *Energies*, vol. 17, no. 15, Art. no. 15, Jan. 2024, doi: 10.3390/en17153731.

[11] O. Melnychenko, L. Scislo, O. Savenko, A. Sachenko, and P. Radiuk, "Intelligent integrated system for fruit detection using multi-UAV imaging and deep learning," *Sensors*, vol. 24, no. 6, p. 1913, Mar. 2024, doi: 10.3390/s24061913.

[12] H. Fesenko, V. Kharchenko, A. Sachenko, R. Hiromoto, and V. Kochan, "An internet of drone-based multi-version post-severe accident monitoring system: Structures and reliability," in *Dependable IoT for Human and Industry*, 1st ed., in Computer Science, Engineering & Technology. New York, NY, USA: River Publishers, 2018, pp. 197–217, doi: 10.1201/9781003337843-12.

[13] B. Alzahrani, O. S. Oubbati, A. Barnawi, M. Atiquzzaman, and D. Alghazzawi, "UAV assistance paradigm: State-of-the-art in applications and challenges," *J. Netw. Comput. Appl.*, vol. 166, p. 102706, Sep. 2020, doi: 10.1016/j.jnca.2020.102706.

[14] O. Melnychenko, O. Savenko, and P. Radiuk, "Apple detection with occlusions using modified YOLOv5-v1," in *2023 IEEE 12th Int. Conf. Intell. Data Acquisition Adv. Comput. Syst.: Technol. Appl. (IDAACS)*, Dortmund, Germany, Sep. 7–9, 2023. New York, NY, USA: IEEE, 2023, pp. 107–112. doi: 10.1109/IDAACS58523.2023.10348779.

[15] V. Lysenko, O. Opryshko, D. Komarchuk, N. Pasichnyk, N. Zaets, and A. Dudnyk, "Information support of the remote nitrogen monitoring system in agricultural crops," *Int. J. Comput.*, vol. 17, no. 1, pp. 47–54, Mar. 2018, doi: 10.47839/ijc.17.1.948.

[16] I. Gohar, J. See, A. Halimi, and W. K. Yew, "Automatic defect detection in wind turbine blade images: Model benchmarks and re-annotations," in *2023 IEEE Int. Conf. Multimedia Expo Workshops (ICMEW)*, Brisbane, Australia, Jul. 10–14, 2023. New York, NY, USA: IEEE Inc., 2023, pp. 290–295. doi: 10.1109/ICMEW59549.2023.00056.

[17] A. ul Husnain, N. Mokhtar, N. Mohamed Shah, M. Dahari, and M. Iwahashi, "A systematic literature review (SLR) on autonomous path planning of unmanned aerial vehicles," *Drones*, vol. 7, no. 2, Art. no. 2, Feb. 2023, doi: 10.3390/drones7020118.

[18] B. Cetinsaya, D. Reiners, and C. Cruz-Neira, "From PID to swarms: A decade of advancements in drone control and path planning - A systematic review (2013–2023)," *Swarm Evol. Computation*, vol. 89, p. 101626, Aug. 2024, doi: 10.1016/j.swevo.2024.101626.

[19] K. Zhang, V. Pakrashi, J. Murphy, and G. Hao, "Inspection of floating offshore wind turbines using multi-rotor unmanned aerial vehicles: Literature review and trends," *Sensors*, vol. 24, no. 3, Art. no. 3, Jan. 2024, doi: 10.3390/s24030911.

[20] M. Memari, P. Shakya, M. Shekaramiz, A. C. Seibi, and M. A. S. Masoum, "Review on the advancements in wind turbine blade inspection: integrating drone and deep learning technologies for enhanced defect detection," *IEEE Access*, vol. 12, pp. 33236–33282, 2024, doi: 10.1109/ACCESS.2024.3371493.

[21] M. Car, L. Markovic, A. Ivanovic, M. Orsag, and S. Bogdan, "Autonomous wind-turbine blade inspection using LiDAR-equipped unmanned aerial vehicle," *IEEE Access*, vol. 8, pp. 131380–131387, 2020, doi: 10.1109/ACCESS.2020.3009738.

[22] D. Pérez, A. Alcántara, and J. Capitán, "Distributed trajectory planning for a formation of aerial vehicles inspecting wind turbines," in *2022 Int. Conf. Unmanned Aircr. Syst. (ICUAS)*, Dubrovnik, Croatia, Jun. 21–24, 2022. New York, NY, USA: IEEE, 2022, pp. 646–654. doi: 10.1109/ICUAS54217.2022.9836173.

[23] A. Shihavuddin *et al.*, "Image based surface damage detection of renewable energy installations using a unified deep learning approach," *Energy Rep.*, vol. 7, pp. 4566–4576, Nov. 2021, doi: 10.1016/j.egyr.2021.07.045.

[24] M. Memari, M. Shekaramiz, M. A. S. Masoum, and A. C. Seibi, "Data fusion and ensemble learning for advanced anomaly detection using multi-spectral RGB and thermal imaging of small wind turbine blades," *Energies*, vol. 17, no. 3, Art. no. 3, Jan. 2024, doi: 10.3390/en17030673.

[25] I. Gohar, A. Halimi, J. See, W. K. Yew, and C. Yang, "Slice-aided defect detection in ultra high-resolution wind turbine blade images," *Machines*, vol. 11, no. 10, Art. no. 10, Oct. 2023, doi: 10.3390/machines11100953.

[26] O. Melnychenko and O. Savenko, "A self-organized automated system to control unmanned aerial vehicles for object detection," in *4th Int. Work. on Intell. Inform. Technol. & Syst. of Inform. Secur. (IntelITSIS)*, Khmelnytskyi, Ukraine, Mar. 22–24, 2023. Aachen, Germany: CEUR-WS.org, 2023, pp. 589–600. [Online]. Available: https://ceur-ws.org/Vol-3373/paper40.pdf

[27] B. Grindley, K. J. Parnell, T. Cherett, J. Scanlan, and K. L. Plant, "Understanding the human factors challenge of handover between levels of automation for uncrewed air systems: a systematic literature review," *Transp. Planning Technol.*, pp. 1–26, 2024, doi: 10.1080/03081060.2024.2375645.

[28] S. Aggarwal and N. Kumar, "Path planning techniques for unmanned aerial vehicles: A review, solutions, and challenges," *Comput. Commun.*, vol. 149, pp. 270–299, Jan. 2020, doi: 10.1016/j.comcom.2019.10.014.

[29] C. Castelar Wembers, J. Pflughaupt, L. Moshagen, M. Kurenkov, T. Lewejohann, and G. Schildbach, "LiDAR-based automated UAV inspection of wind turbine rotor blades," *J. Field Robot.*, vol. 41, no. 4, pp. 1116–1132, 2024, doi: 10.1002/rob.22309.

[30] Z. Li, J. Wu, J. Xiong, and B. Liu, "Research on automatic path planning of wind turbines inspection based on combined UAV," in *2024 IEEE Int. Symp. Broadband Multimedia Syst. Broadcast. (BMSB)*, Toronto, ON, Canada, Jun. 19–21, 2024. New York, NY, USA: IEEE, Jun. 2024, pp. 1–6. doi: 10.1109/BMSB62888.2024.10608306.

[31] H. Liu, Y. P. Tsang, C. K. M. Lee, and C. H. Wu, "UAV trajectory planning via viewpoint resampling for autonomous remote inspection of industrial facilities," *EEE Trans. Ind. Inform.*, vol. 20, no. 5, pp. 7492–7501, May 2024, doi: 10.1109/TII.2024.3361019.

[32] facebookresearch/detectron2. (Jul. 17, 2024). Python. Meta Research. Accessed: Aug. 21, 2024. [Online]. Available: https://github.com/facebookresearch/detectron2

[33] D. Zahorodnia, Y. Pigovsky, and P. Bykovyy, "Canny-based method of image contour segmentation," *Int. J. Comput.*, vol. 15, no. 3, pp. 200–205, Sep. 2016, doi: 10.47839/ijc.15.3.853.

[34] S. Gollapudi, "OpenCV with Python," in *Learn Computer Vision Using OpenCV*, S. Gollapudi, Ed., Berkeley, CA: Apress, 2019, pp. 31–50. doi: 10.1007/978-1-4842-4261-2_2.

[35] I. Obeidat and M. AlZubi, "Developing a faster pattern matching algorithms for intrusion detection system," *Int. J. Comput.*, vol. 18, no. 3, pp. 278–284, Sep. 2019, doi: 10.47839/ijc.18.3.1520.

[36] I. Lopez-Sanchez and J. Moreno-Valenzuela, "PID control of quadrotor UAVs: A survey," *Annu. Rev. Control*, vol. 56, p. 100900, Jan. 2023, doi: 10.1016/j.arcontrol.2023.100900.